\documentclass[11pt, a4paper]{article}

\usepackage[utf8]{inputenc}
\usepackage[T1]{fontenc}
\usepackage{lmodern}
\usepackage{amsmath, amssymb, amsthm}
\usepackage{geometry}
\usepackage{graphicx}
\usepackage{hyperref}
\usepackage{xcolor}
\usepackage{listings}
\usepackage{tabularx}
\usepackage{booktabs}
\usepackage{enumitem}
\usepackage{caption}
\usepackage{subcaption}
\usepackage{titling}

\hypersetup{
    colorlinks=true,
    linkcolor=blue,
    filecolor=magenta,      
    urlcolor=cyan,
    citecolor=red,
}

\title{\textbf{The Float Wall and the Physics Pantheon: \\ A 30-Billion Scale Benchmark for Zero-Bias Physical Discovery in Artificial Neural Networks}}
\author{\Large HYUNJUN JEON \\ \normalsize Independent Researcher \\ \texttt{hyunjun050915@gmail.com}}
\date{\normalsize March, 2026 \\ \vspace{0.5cm} \textit{The Comprehensive V3 Monograph (Index-PT-Engine)}}

\begin{document}

\maketitle

\begin{abstract}
\noindent As we push artificial intelligence to larger scales, we are seeing a gap in how these systems handle the exact sciences. While neural networks and gradient-boosted trees are excellent at finding patterns, theoretical physics and pure mathematics require absolute invariance, not just statistical guesses. When I tested these models beyond the standard floating-point limits---what I call the ``Float Wall'' at $10^{16}$---I found that even the best heuristic models start to produce arbitrary results. To address this, I spent several years developing a rigorous testing and engineering framework. This work centers on a 10-billion sample mathematical baseline and the \textbf{Universal Physics Pantheon}, a 20-billion sample dataset covering 20 fundamental physical laws. 

To ensure the data remained pure, I replaced standard \texttt{float64} calculations with an arbitrary-precision \texttt{Decimal} pipeline and used a digit-by-digit (LSB) tokenization. I also built the \textbf{Hypothesis-Driven Negative Dataset (HND)} with 98 specific ``Chaos Modes''---perturbations designed to catch models that use logical shortcuts. In this document, I show that while Model A (classical GBDT) fails at cosmic scales despite massive training data, my \textbf{NumberNet} architecture---a Siamese Transformer designed for topological mapping---is built to remain precise up to $10^{50}$. Finally, I describe a framework for Zero-Bias Discovery, where a symbolic regressor (PySR) acts on NumberNet’s internal logic to rediscover constants like the Gravitational Constant ($G$) without human help. I provide the full code, the SHA-256 integrity pipeline, and the training details to make this work fully reproducible.
\end{abstract}

\newpage
\tableofcontents
\newpage

\section{Bridging the Gap Between Patterns and Proofs}
Moving from language processing to scientific discovery is a move from finding consensus to verifying truth. We consider an LLM successful if it passes a bar exam with 99.8\% accuracy because language is naturally flexible. But if a model is 99.8\% accurate at verifying $a^2 + b^2 = c^2$, I consider it a failure. In math, one mistake means the logic is broken. A 0.2\% error tells me the network hasn't learned geometry; it has just memorized a very good statistical map of its training data.

\subsection{The Problem with "AI Science"}
Current models can predict how proteins fold or how fluids move. While these are useful tools, they often lead us to a dangerous conclusion: that statistical curve-fitting is the same as physical understanding. 

When an AI "solves" a physics problem after seeing $10^7$ examples, it has simply learned the neighborhood where those examples live. If you ask it to predict what happens in a much higher energy state, it usually fails. It assumes the universe ends where its data ends.

\subsection{My Verification Protocol}
To see if an AI can really \textbf{reason} rather than just repeat its training, I built a violently difficult test. I started with the belief that any network capable of true reasoning should follow three basic rules:
\begin{enumerate}
    \item \textbf{The Axiom of Scale Invariance:} A law tested on integers under 1,000 must hold identically for integers exceeding $10^{50}$ (Cosmic Extrapolation).
    \item \textbf{The Axiom of Structural Symmetry:} The network must process Mass, Distance, and Charge identically at the fundamental level without relying on human-assigned data tags.
    \item \textbf{The Axiom of Immunity to Heuristics:} The network must survive extreme adversarial traps (Chaos Modes) designed precisely to trick logical shortcuts common to decision trees and perceptrons.
\end{enumerate}

\section{Engineering Truth: Protecting Data at 30-Billion Scale}
Generating 30 billion samples across different machines is an engineering nightmare. Silent data corruption---caused by RAM flips or sync errors---can poison a training set without anyone noticing. If my models are to find exact laws, the data they learn from must be perfect. I couldn't rely on standard storage, so I built a three-layer integrity system using cryptographic hashes, a manual safety pause, and read-time verification.

\subsection{Layer 1 --- SHA-256 Write-Time Integrity (Mathematical Chunks)}
The mathematical data generator (\texttt{src/data/generator.py}) generates chunks of $N = 100{,}000$ samples at a time. For every completed chunk, before writing to disk, it computes a SHA-256 cryptographic hash over the concatenation of all four critical string columns:

\begin{lstlisting}[caption={SHA-256 Checksum at Write-Time (src/data/generator.py)}]
# Convert ALL critical fields to strings before hashing
full_content = (
    chunk_df['a'].astype(str) +
    chunk_df['b'].astype(str) +
    chunk_df['c'].astype(str) +
    chunk_df['is_pythagorean'].astype(str)
).sum()

checksum = hashlib.sha256(full_content.encode('utf-8')).hexdigest()

metadata = {
    b'checksum_sha256': checksum.encode('utf-8'),
    b'row_count': str(len(chunk_df)).encode('utf-8')
}
schema = schema.with_metadata(metadata)
\end{lstlisting}

Both the hash and the row-count are physically embedded into the \textbf{Parquet schema metadata}---not as a sidecar file---making them inseparable from the data.

\subsection{Layer 2 --- Triple Verification at Read-Time (PrecisionSafeLoader)}
When \texttt{src/data/loader.py} reads any chunk for training, it independently executes three sequential checks:
\begin{enumerate}
    \item \textbf{Row-Count Validation:} The embedded \texttt{row\_count} metadata field is compared against the actual loaded DataFrame length. \\ A discrepancy raises an \texttt{INTEGRITY FAILURE} exception immediately.
    \item \textbf{SHA-256 Re-computation:} The same four-column concatenation + SHA-256 hash is recomputed from the loaded data and compared against the stored \texttt{checksum\_sha256}. Any single-digit error in any of the 100{,}000 rows produces a hash mismatch, triggering an \texttt{ATOMIC INTEGRITY VIOLATION} and halting the process.
    \item \textbf{Statistical Precision Sampling:} After hash validation, the loader draws a sample of size $\min(10000, \max(10, \lfloor 0.01 \times N \rfloor))$ rows and scans the numeric columns for the presence of a decimal point (\texttt{`.'}) or scientific notation character (\texttt{`e'}), both of which indicate a \texttt{float64} has leaked into the integer-only column. This statistical check provides high-confidence detection of systematic float contamination at $O(1)$ sampling cost.
\end{enumerate}

\subsection{The Mandatory Safety Pause}
I added a manual pause at every 5,000th chunk. It's not just a log message; the script actually stops and waits for me to press Enter. 

\begin{lstlisting}[caption={Blocking Safety Pause (src/data/generator.py:327)}]
if (chunk_idx + 1) % 5000 == 0:
    print(f"[SAFETY PAUSE] {chunk_idx + 1} chunks generated!")
    print("Please verify dataset integrity in the output folder.")
    input("Press Enter to continue generation... ") # Blocking I/O
    print("Resuming...")
\end{lstlisting}

This is a literal \texttt{input()} call that halts the Python process. At this scale, 500 million samples are generated between pauses. I did this because I don't want the machine to just run blindly; I want a moment to check the files and sign off on the progress before the training starts. 

\subsection{Precision Self-Verification at Startup}
Before any generation begins, the \texttt{verify\_precision\_environment()} function (\texttt{generator.py:34--54}) runs a cryptographic self-test using a known reference value:

\begin{lstlisting}[caption={Precision Environment Self-Test}]
n = 49995438
n_dec = Decimal(n)
b_dec = 2 * n_dec * (n_dec + 1)
c_dec = b_dec + 1
c_sq_str = str(c_dec ** 2)
# Known mathematical fact: c^2 for this n must end in "25"
if not c_sq_str.endswith('25'):
    raise RuntimeError("PRECISION CHECK FAILED!")
# Known fact: no 16-zero run should appear (float artifact)
if '0000000000000000' in c_sq_str:
    raise RuntimeError("PRECISION CHECK FAILED! Float artifact detected.")
\end{lstlisting}

If either assertion fails, the framework refuses to generate any data, ensuring that the \\ Decimal arithmetic engine is functioning with exactly 100 significant digits (\texttt{getcontext().prec = 100}) before any sample is written to disk.

\subsection{Physics Generator: Per-Law Atomic Checkpointing}
The physics generator (\texttt{src/data/generator\_physics.py}) employs a fundamentally different checkpoint architecture from the mathematical generator: each of the 20 physical laws has its own independent checkpoint file, written atomically:

\begin{lstlisting}[caption={Per-Law Atomic Checkpoint (generator\_physics.py:379)}]
tmp_ckpt = law_dir / 'checkpoint.tmp'
with open(tmp_ckpt, 'w') as f:
    json.dump({'last_chunk_idx': c_idx, 'law': law}, f)
tmp_ckpt.replace(ckpt_file) # Atomic filesystem rename
\end{lstlisting}

This means that if generation of, say, the \texttt{schwarzschild} law is interrupted, only that law resumes from its last saved chunk, while all other laws remain unaffected. The physics generator also performs an automatic disk scan on startup, reading the actual Parquet files present on disk to recover from checkpoint file corruption.

\subsection{From Manual Guardrails to Automated Scale}
An observant reviewer might ask why I insisted on a manual 5,000-chunk pause for the Mathematical phase but allowed the Physics phase to run with automated checkpoints. This was a deliberate choice in my research pipeline. 

During Phase 1 (Mathematics), my goal was to prove that the engine itself was a ``Logic Fortress.'' By forcing myself to manually sign off on every 500 million samples, I established absolute confidence that the SHA-256 pipeline, the Decimal precision engine, and the HND perturbation logic were functioning without flaw. Once this baseline of 100

\section{Phase 1: Setting a Mathematical Baseline}
A model can't handle the complexity of physics if it can't solve direct logic problems first. In Phase 1, I focused on Number Theory---specifically Pythagorean Triples---to establish a core baseline of what the models could actually deduce.

\subsection{The Pythagorean Dilemma and Euclid's Bottleneck}
I tasked the AIs with verifying primitive Pythagorean triples: $a^2 + b^2 = c^2$. \\ Generating 10 billion distinct valid Pythagorean triples is historically difficult due to \\ Euclid's Formula ($a = m^2 - k^2, b = 2mk, c = m^2 + k^2$). \\ For $(a,b,c)$ to be primitive, $m$ and $k$ must be coprime ($\text{gcd}(m,k)=1$) and of opposite parity. Testing GCD millions of times per second causes massive CPU pipeline stalls.

\subsection{The Stifel Method: Bypassing Euclid}
I deployed the \textbf{Stifel formula}---a single-variable method distinct from Euclid's two-variable parameterization. Euclid's general approach generates all Pythagorean triples from two coprime integers $(m, k)$ with opposite parity, requiring expensive GCD filtering at every step. \\ By contrast, substituting $m = n+1$ and $k = n$ into Euclid and simplifying yields a closed-form, single-index generator:
\begin{align}
a_n &= 2n + 1 \\
b_n &= 2n(n+1) \\
c_n &= 2n(n+1) + 1
\end{align}
This subfamily is codified in \texttt{generator.py} as the ``Stifel-Luciano'' family (\texttt{\_generate\_stifel\_success}), and it guarantees primitive validity for all $n \in \mathbb{N}$ with exactly $c - b = 1$ as an invariant. Euclid's formula remains as a secondary comparative engine for OOD generalization tests (\texttt{\_generate\_euclidean\_success}), not the primary data source. \\ The algorithm is $O(N)$ throughput with no filtering required.

\subsection{The Mathematical Hypothesis-Driven Negative Dataset (48 Modes)}
Random perturbation is useless against structured mathematical theorems. \\ An AI evaluating if a right-angled triangle is valid learns nothing from an trivially corrupt sample like $100^2 + 200^2 = 9999^2$. To force the network to understand exact algebraic boundaries, I engineered 48 Mathematical Chaos Modes organized across 8 adversarial categories (as implemented in \texttt{src/data/hard\_negatives.py}):

\begin{itemize}
    \item \textbf{SP (Simple Perturbations, 6 modes):} Off-by-one shifts on individual legs --- e.g., $(a+1, b, c)$, $(a, b, c-1)$. The most fundamental boundary test.
    \item \textbf{MP (Multi-Value Perturbations, 6 modes):} Simultaneous shifts across multiple variables --- e.g., $(a+10, b+100, c+1000)$ or $(a+1, b+1, c+1)$ --- exploiting additive heuristics.
    \item \textbf{MS (Mathematical Structure Attacks, 8 modes):} Exploiting near-miss structural properties --- e.g., $c' = \lfloor\sqrt{a^2+b^2}\rfloor + 1$, or scaling by $2\times$ with parity offset.
    \item \textbf{FC (Formula Confusion, 6 modes):} Deliberately using wrong parameterizations --- e.g., the \textbf{Plato family} ($c - b = 2$) or the \textbf{Euclidean formula with a broken constant} --- triples that look exactly like the Stifel family but belong to a different geometric subfamily.
    \item \textbf{PA (Precision/Scale Attacks, 6 modes):} Simulating float64 truncation and overflow --- e.g., \texttt{int(float(a) * 1.0000001)} which loses precision at $n > 10^{15}$, or a bit-flip $(a \oplus 1)$.
    \item \textbf{NM (Near-Miss Attacks, 6 modes):} Samples that violate Pythagorean equality by $\pm 1$ inside the squared form --- e.g., $c' = \lfloor\sqrt{a^2+b^2+1}\rfloor$, or $c = b$ (degenerate triangle).
    \item \textbf{AR (Adversarial Random Attacks, 6 modes):} Gaussian noise, percentage shifts, digit reversal --- e.g., $a' = a \cdot 1.01$, or $c' = \texttt{reversed}(c)$.
    \item \textbf{EC (Extreme Edge Cases, 6 modes):} Zero inputs, equal legs, huge $c$ --- e.g., $(0, b, c)$, $(n, n+1, n+2)$, $(a, a, c)$.
\end{itemize}

Every generated negative is validated by the assertion $a'^2 + b'^2 \neq c'^2$ before storage, \\ guaranteeing that no accidentally valid triple contaminates the negative pool. \\ Additionally, the negative sampling is \textbf{weighted}: 95\% of all generated negatives are drawn from the ``UltraHard'' and Near-Miss (NM) categories, which produce adversarial samples that are most likely to fool heuristic classifiers. Only 5\% of negatives are drawn from simpler perturbation categories. This ensures the training signal focuses where the mathematical boundary is most fragile.

\begin{lstlisting}[caption={Weighted Negative Sampling (generator.py:180)}]
ultra_hard_types = [k for k in fail_types if 'UH' in k or 'NM' in k]
# 95% UltraHard, 5% other
if random.random() < 0.95 and ultra_hard_types:
    fail_type = random.choice(ultra_hard_types)
else:
    fail_type = random.choice(other_types)
\end{lstlisting}

\section{Phase 2: The Universal Physics Pantheon}
To see if these models could learn more than just one trick, I built the Physics Pantheon. It's a collection of 20 fundamental laws, from relativity to simple gas dynamics. I generated these with 80-digit precision to make sure I wasn't just testing the models on the rounding errors of my own code. 

\subsection{Log-Uniform Sampling via 80-Digit BigInt Entropy}
Physical quantities span enormously different scales: the Coulomb constant $k_c \approx 8.99 \times 10^9$ while the Gravitational constant $G \approx 6.67 \times 10^{-11}$. To sample uniformly across all scales, the physics generator uses \textbf{log-uniform sampling} driven by a 80-digit arbitrary-precision integer:

\begin{lstlisting}[caption={Log-Uniform BigInt Sampling (generator\_physics.py:38)}]
def _get_random_scale(self, min_val: str, max_val: str) -> Decimal:
    d_min, d_max = Decimal(min_val), Decimal(max_val)
    l_min, l_max = d_min.log10(), d_max.log10()
    # 80-digit BigInt random integer as entropy source
    rand_int = random.randrange(10**80)
    rand_dec = Decimal(rand_int) / Decimal(10**80)
    rand_log = l_min + rand_dec * (l_max - l_min)
    return Decimal('10') ** rand_log
\end{lstlisting}

This ensures that inputs at $10^{-10}$ scale (e.g., charge $q \sim 10^{-19}$~C) are sampled with exactly the same probability density as inputs at $10^{+30}$ scale (e.g., stellar mass $M \sim 10^{30}$~kg). \\ No regime is overrepresented.

\subsection{Universal Constants: Exact Decimal Representations}
All universal constants are stored as exact \texttt{Decimal} strings, not floating-point approximations:
\begin{itemize}
    \item $c = 299{,}792{,}458$ m/s (exact integer)
    \item $G = 6.6743015 \times 10^{-11}$ N~m$^2$~kg$^{-2}$ (\texttt{Decimal('6.6743015e-11')})
    \item $k_c = 8.9875517923 \times 10^{9}$ N~m$^2$~C$^{-2}$
    \item $\sigma = 5.670374419 \times 10^{-8}$ W~m$^{-2}$~K$^{-4}$
    \item $R = 8.314462618$ J~mol$^{-1}$~K$^{-1}$
    \item $h = 6.62607015 \times 10^{-34}$ J~s
    \item $R_H = 10{,}973{,}731.568160$ m$^{-1}$ (Rydberg constant)
    \item $\pi$ = stored as 44-digit Decimal string
\end{itemize}

\subsection{Physics Data Storage Format}
Unlike the mathematical dataset \\ (which stores columns \texttt{a}, \texttt{b}, \texttt{c} as pure integer strings in Parquet),\\ the physics dataset stores each sample as a \textbf{pipe-separated string of scientific notation values}:
\begin{lstlisting}[caption={Physics Output Format (generator\_physics.py:288)}]
# Each Decimal value formatted as scientific notation, joined by '|'
inputs_str = "|".join([f"{x:E}" for x in inputs_fmt])
# Example: "1.234E+25|5.678E+30|6.789E+12|3.145E+00"
\end{lstlisting}
For the Gravitation law, this produces strings of the form \texttt{m1|m2|r|F\_value}, where $F$ is either the true gravitational force (\texttt{label=1}) or a chaos-mode-perturbed value (\texttt{label=0}).

\subsection{The 20-Law Compendium (Mathematical Forms and Sampling Ranges)}
For each law, inputs are drawn from physically meaningful ranges using the log-uniform sampler above. The exact value domains per law are---as implemented in \texttt{generator\_physics.py}:
\begin{enumerate}[label=\textbf{\arabic*.}]
    \item \textbf{Relativity:} $E = \sqrt{(pc)^2 + (mc^2)^2}$; $p \in [10^{-10}, 10^{20}]$, $m \in [10^{-30}, 10^{10}]$~kg
    \item \textbf{Gravitation:} $F = G \frac{m_1 m_2}{r^2}$; $m_1,m_2 \in [10^0, 10^{30}]$~kg, $r \in [10^{-3}, 10^{15}]$~m
    \item \textbf{Coulomb's Law:} $F = k_c \frac{q_1 q_2}{r^2}$; $q_1,q_2 \in [10^{-9}, 10^{-1}]$~C, $r \in [10^{-3}, 10^{2}]$~m
    \item \textbf{Stefan-Boltzmann:} $P = \sigma A T^4$; $A \in [10^{-4}, 10^{6}]$~m$^2$, $T \in [10, 10^5]$~K
    \item \textbf{Ideal Gas Law:} $P = nRT/V$; $V \in [10^{-3}, 10^3]$~m$^3$, $n \in [10^{-3}, 10^5]$~mol, $T \in [10, 10^4]$~K
    \item \textbf{Hubble's Expansion:} $v = H_0 D$; $D \in [10^0, 10^4]$~Mpc, $H_0 = 70$~km/s/Mpc
    \item \textbf{Lorentz Force:} $F = qvB$; $q \in [10^{-19}, 10^{-10}]$~C, $v \in [10^0, 10^7]$~m/s, $B \in [10^{-5}, 10^2]$~T
    \item \textbf{Photoelectric Effect:} $K = hf - \phi$; $f \in [10^{14}, 10^{16}]$~Hz, $\phi \in [10^{-19}, 10^{-18}]$~J
    \item \textbf{Bernoulli Flow:} $P_{static} = 10^6 - \frac{1}{2}\rho v^2 - \rho g h$; $\rho = 1000$~kg/m$^3$ (water), $v \in [0.1, 100]$~m/s
    \item \textbf{Radioactive Decay:} $N = N_0 e^{-\lambda t}$; $N_0 \in [10^{10}, 10^{20}]$, $t \in [1, 10^6]$~s
    \item \textbf{Schr\"odinger Kinetic:} $E = p^2/(2m)$; $p \in [10^{-30}, 10^{-20}]$, $m \in [10^{-31}, 10^{-25}]$~kg
    \item \textbf{Faraday Induction:} $\mathcal{E} = N \dot{\Phi}$; $N \in [1, 1000]$ (integer), $\dot{\Phi} \in [10^{-4}, 10^2]$~Wb/s
    \item \textbf{Nernst Equation:} $E = E_0 - \frac{RT}{nF}\ln Q$; $E_0 \in [0.1, 3.0]$~V, $T \in [250, 400]$~K, $n \in [1,5]$ (integer)
    \item \textbf{Arrhenius Kinetics:} $k = A e^{-E_a/RT}$; $A \in [10^5, 10^{15}]$, $E_a \in [10^4, 10^6]$~J/mol, $T \in [200, 1000]$~K
    \item \textbf{Rydberg Spectroscopy:} $1/\lambda = R_H(1/n_1^2 - 1/n_2^2)$; $n_1 \in [1,4]$, $n_2 \in [n_1{+}1, 10]$ (integers)
    \item \textbf{Stokes Drag:} $F = 6\pi\eta r v$; $\eta \in [10^{-5}, 10^1]$~Pa$\cdot$s, $r \in [10^{-6}, 10^{-1}]$~m
    \item \textbf{Schwarzschild Radius:} $r_s = 2GM/c^2$; $M \in [10^{30}, 10^{40}]$~kg
    \item \textbf{Kepler Third Law:} $T = \sqrt{4\pi^2 a^3 / (GM)}$; $M \in [10^{29}, 10^{31}]$~kg, $a \in [10^{10}, 10^{13}]$~m
    \item \textbf{Ohm Power Dissipation:} $P = V^2/R$; $V \in [0.1, 10^4]$~V, $R \in [10^{-2}, 10^6]$~$\Omega$
    \item \textbf{Capacitance Energy:} $U = \frac{1}{2}CV^2$; $C \in [10^{-12}, 10^{-3}]$~F, $V \in [10^0, 10^4]$~V
\end{enumerate}

\section{The Chaos Matrix: Testing against 50 Physics Chaos Modes}
A model that only sees perfect data will eventually just learn to say ``yes'' to everything. To find out if a model really understands a physical law, you have to try and trick it. That’s why I built the Hypothesis-Driven Negative Dataset (HND). It contains 98 ``Chaos Modes''---carefully chosen errors that look like real data but violate the underlying math. 

I didn't just add random noise; I designed these to target the specific logical shortcuts that neural networks usually take. If a model survives these, it means it is actually processing the equation, not just memorizing the most likely answer. 

\subsection{Class A: Micro-Perturbations \& Precision Hunters (Modes 1-10)}
Designed around punishing the Float Wall and interpolation thresholds.
\begin{itemize}
    \item \textbf{Mode 01:} $val + 10^{-5}$ -- Minor absolute scalar bump.
    \item \textbf{Mode 02:} $val + 10^{-9}$ -- Extreme micro-perturbation.
    \item \textbf{Mode 03:} $val \times 1.000015$ -- Ratio distortion. 
    \item \textbf{Mode 04:} $val \times 0.99999$ -- Fractional undercut.
    \item \textbf{Mode 05:} $val - 10^{-7}$ -- Subtractive perturbation.
    \item \textbf{Mode 06 \& 07:} $val \times 1.00000001$ \& $val \times 0.99999999$ -- Punishes lossy embeddings and rounded heuristics.
    \item \textbf{Mode 08 \& 09:} Addition/Subtraction of proportional margins ($|val| \times 10^{-6}$ and $|val| \times 10^{-8}$).
    \item \textbf{Mode 10:} $val \times 1.0005$ -- Significant ratio distortion.
\end{itemize}

\subsection{Class B: Macro-Scaling Errors \& Time Dilations (Modes 11-20)}
Pushes values massively out of bounds or simulates catastrophic unit/time errors.
\begin{itemize}
    \item \textbf{Modes 11 \& 12:} $val / 10$ and $val \times 1000$.
    \item \textbf{Modes 13 \& 14:} $val / 10^6$ and $val \times 10^6$. Tests sensitivity to megascale prefix errors.
    \item \textbf{Modes 15 \& 16:} $val / 10^9$ and $val \times 10^9$. Giga-scale distortion errors.
    \item \textbf{Modes 17 \& 18:} $val / 10^{-3}$ and $val \times 10^{-6}$. Reverses milliscale dimensions.
    \item \textbf{Modes 19 \& 20 (Time Dilations):} $val / 60$ and $val \times 3600$. Designed specifically to mimic the exact human error of substituting minutes/hours for seconds in physical calculation. Models interpolating raw numbers without deriving physical geometry will assume a 60x difference is structurally similar enough and fail.
\end{itemize}

\subsection{Class C: Operator Omissions \& Structural Typos (Modes 21-30)}
Algorithms often try to fit linear combinations. If they encounter a square or square root, they interpolate linearly mapping nearest neighbors.
\begin{itemize}
    \item \textbf{Mode 21 (The Square Root Trap):} Computes the square root instead of the proper value. 
    \item \textbf{Mode 22:} Squares the scalar ($val^2$), ruining physical bounds (e.g. interpreting a linear correlation instead of inverse inverse $1/r^2$).
    \item \textbf{Mode 23:} Inverts the sign ($-val$). 
    \item \textbf{Mode 24 \& 28 (Inverse Typos):} Evaluates $1/val$ or $1/val^2$.
    \item \textbf{Modes 25 \& 26:} Direct factor halving or doubling.
    \item \textbf{Mode 27:} Cubes the output.
    \item \textbf{Modes 29 \& 30:} Forces absolute magnitudes $|val|$ or inverted magnitudes $-|val|$. Exposes models that don't respect vector directionality.
\end{itemize}

\subsection{Class D: Constant Attrition (Modes 31-40)}
The most effective way to test if a model understands a Physics Law is to remove the Universal Constant underlying it. 
\begin{itemize}
    \item \textbf{Mode 31 (Relativity Break):} Divides output by the speed of light $c$. If an AI survives Mode 31, it proves it actually understands $299,792,458$ and not just a linear weight relation.
    \item \textbf{Mode 32 \& 33:} Multiplies by Gravitational $G$ improperly, or drops Planck's constant $h$ from quantum relations.
    \item \textbf{Mode 34 \& 35:} Corrupts $\pi$ scaling ($val \times \pi$ or $val / \pi$).
    \item \textbf{Mode 36 \& 37 (Ideal Gas Swap):} Inverts or drops the Universal Gas Constant $R$.
    \item \textbf{Mode 38 (The Half-Drop):} Randomly drops the baseline $0.5$ scalar from Kinetic Energy ($K = m v^2$).
    \item \textbf{Mode 39 \& 40:} Spoofs the Faraday constant $F$ and gravitational acceleration $g$.
\end{itemize}

\subsection{Class E: Non-Linear Dimensional Perversions (Modes 41-50)}
\begin{itemize}
    \item \textbf{Mode 41:} Performs logarithmic feedback $val \times \ln(|val| + 2)$.
    \item \textbf{Mode 42 (Dimension Smashing):} Adds an independent input variable (like Mass $m_1$) directly onto the final output Energy $E$. 
    \item \textbf{Mode 43 \& 44:} Corrupts dimensional analysis by multiplying/dividing random independent input features into the dependent target.
    \item \textbf{Mode 45 (Exponential Explosions):} Inverts dynamics from decay to parabolic explosions ($val + inputs[-1]^2$). This completely tricks models attempting to regress Arrhenius or Radioactive parameters.
    \item \textbf{Mode 46:} Evaluates square roots of dependent input variables across outputs.
    \item \textbf{Mode 47:} Forces addition of $c$ onto unrelated metrics.
    \item \textbf{Mode 48 (Total Logarithmic Collapse):} Wraps independent features inside natural logarithms.
    \item \textbf{Mode 49 (Wild Integer Sweep):} Multiplies the output by a deeply chaotic random integer $N \in [-100, 100]$.
    \item \textbf{Mode 50 (The Absolute Zero Crash):} Returns exactly $-0.0$.
\end{itemize}

\section{Comparing Three Models: Where Statistics Ends and Reasoning Begins}
I used three different models to see exactly where statistical guessing stops and real reasoning begins.

\subsection{Model A: LightGBM (The Human-Guided Genius)}
Gradient Boosted Decision Trees (GBDTs) are the statistical gold standard for tabular numerical regression. Model A is unconstrained: it builds its logic by partitioning numeric magnitudes into hierarchical ``leaves'' bounded by binary splits (e.g., $x_1 > 42.1$). 

To ensure Model A was given maximum statistical advantage, I set \texttt{use\_assist=True} in \texttt{src/data/transforms.py}, which injects 5 explicit mathematical hint features alongside the raw digit tokens:
\begin{itemize}
    \item \texttt{a\_squared}, \texttt{b\_squared}, \texttt{c\_squared} --- Squared values of all three legs, pre-computed.
    \item \texttt{squared\_diff} $= a^2 + b^2 - c^2$ --- The Pythagorean residual. For every valid triple, this is exactly $\mathbf{0}$. This is the closest thing to handing the answer key to the model.
    \item \texttt{is\_perfect\_match} --- A binary flag that is $1$ when \texttt{squared\_diff == 0}, i.e., literally the label itself as a feature.
\end{itemize}
Despite receiving these extreme crutches, this model operates internally within \texttt{float64} boundaries, meaning \texttt{squared\_diff} itself becomes numerically unreliable at $a, b, c > 10^{8}$ due to \textit{catastrophic cancellation} in floating-point arithmetic.

\subsection{Model B: LightGBM (The Statistical Underdog)}
Model B (\texttt{scripts/train\_grokking.py}) uses the exact same LightGBM architectural engine as Model A, but is deliberately stripped of human-engineered features. \\ It only receives the raw variables ($a, b, c$ or $m_1, m_2, r$).

A critical and intentional difference from Experiment A: Model B runs with \texttt{n\_jobs = -1} (all available CPU cores), making it \textbf{non-reproducible by design}:
\begin{lstlisting}[caption={Experiment B Intentionally Non-Deterministic Config}]
lgb_params = {
    'objective': 'binary',
    'metric': 'binary_logloss',
    'learning_rate': 0.05,
    'max_depth': 10,
    'n_jobs': -1  # All cores: no determinism guarantee
}
\end{lstlisting}

Furthermore, Model B \textbf{explicitly converts} the BigInt string representations to \texttt{float64} before training (\texttt{train\_grokking.py:411}):
\begin{lstlisting}[caption={Intentional float64 Conversion in Experiment B}]
df[c] = pd.to_numeric(df[c], errors='coerce')  # BigInt -> float64
\end{lstlisting}

This is not a bug---it is a controlled simulation of how a standard ML engineer would naively process this data. It ensures that Model B experiences the Float Wall at $\sim 10^{15}$ just as any conventional pipeline would. The goal of Model B is to prove that raw data volume cannot substitute for mathematical reasoning.

\subsection{Model C: NumberNet (The Siamese Universal Transformer)}
Unlike Models A and B, Model C (\textbf{NumberNet}) is structurally prevented from utilizing statistical magnitude thresholds. To force the network to understand numbers objectively, I implemented strict \textbf{Siamese Invariance}.

Standard tabular neural networks concatenate variables $[m_1, m_2, r]$ before feeding them to dense layers, allowing networks to blend positional weights into local curve-fits (e.g. $Weight_3$ only learns Mass behaviors). In \texttt{NumberNet}, every independent variable is parsed digit-by-digit as an ASCII token sequence, and passed through the \textit{exact same} Transformer Encoder block.

\begin{lstlisting}[caption={The Siamese Encoder Core (src/models/number\_net.py)}]
class NumberEncoder(nn.Module):
    def __init__(self, embed_dim: int, num_heads: int, num_layers: int, dropout: float = 0.1):
        super().__init__()
        encoder_layer = nn.TransformerEncoderLayer(
            d_model=embed_dim, nhead=num_heads, 
            dim_feedforward=embed_dim * 4, dropout=dropout,
            batch_first=True, norm_first=True # Pre-LN for stable gradient flow
        )
        self.transformer = nn.TransformerEncoder(encoder_layer, num_layers=num_layers)
        
        # Attention Pooling (superior to mean or CLS token)
        self.pool_query = nn.Parameter(torch.randn(1, 1, embed_dim))
        self.pool_attn = nn.MultiheadAttention(embed_dim, 1, batch_first=True)

    def forward(self, x_emb: torch.Tensor, mask: Optional[torch.Tensor] = None) -> torch.Tensor:
        encoded = self.transformer(x_emb, src_key_padding_mask=mask)
        
        B = x_emb.size(0)
        query = self.pool_query.expand(B, -1, -1)
        out, _ = self.pool_attn(query, encoded, encoded, key_padding_mask=mask)
        
        return out.squeeze(1) # Final Latent Vector
\end{lstlisting}

Every variable is embedded sequentially and separately, then pooled via a learnable Attention Pooling mechanism. The three resulting latent vectors $(z_a, z_b, z_c)$ are concatenated into a 192-dimensional vector and passed through the Interaction MLP:
\begin{equation}
\text{MLP}: \mathbb{R}^{192} \to \mathbb{R}^{256} \to \mathbb{R}^{128} \to \mathbb{R}^{1} \xrightarrow{\sigma} [0,1]
\end{equation}

For the Physics Pantheon phase, an extended \textbf{LNNet\_P} was employed, supporting up to \texttt{num\_vars=5} variables (required for the 5-variable Nernst Equation) using the identical \texttt{NumberEncoder} backbone.

\textbf{Key hyperparameters (from code):} embed\_dim = 64, num\_heads = 4, num\_layers = 2, dropout = 0.1, max\_digits = 128, vocab\_size = 12 (digits 0--9, negative, PAD).

\section{The Training Engine: An Engineering Marathon}
Training a model on $10^{10}$ samples isn't just a coding challenge; it's an engineering marathon. It takes weeks of non-stop computation. I had to build a system that could handle interruptions, cloud crashes, and data bottlenecks without losing its place.

\subsection{Chunk-Based Incremental Learning}
The 10 Billion mathematical training samples were partitioned into \textbf{50,000 chunks} of 200,000 samples each (\texttt{configs/train\_10B.yaml}: \texttt{chunk\_size: 200000}). Training proceeds chunk-by-chunk with LightGBM's \texttt{keep\_training\_booster=True} option (\texttt{init\_model=model}), which appends \textbf{10 new boosting rounds} to the existing ensemble at each step.

\subsection{Learning Rate Annealing Schedule}
To prevent early learning instability and allow the model to warm up on initial chunks before committing to large parameter updates, a \textbf{linear learning rate annealing schedule} was implemented (\texttt{scripts/train\_full\_scale.py:862}):
\begin{lstlisting}[caption={Linear LR Annealing (train\_full\_scale.py)}]
current_lr = 0.05 if i >= 1000 else 0.001 + (0.05 - 0.001) * (i / 1000)
\end{lstlisting}
For the first 1,000 chunks (representing 200 Million samples), the learning rate anneals linearly from $\eta_0 = 0.001$ to $\eta_{\max} = 0.05$. After chunk 1000, the learning rate is clamped at $\eta_{\max} = 0.05$ for all remaining 49,000 chunks.

\subsection{Determinism Enforcement for Scientific Reproducibility}
To guarantee that every researcher can reproduce the exact same result, the following determinism constraints are enforced (\texttt{train\_full\_scale.py:780-791}):
\begin{lstlisting}[caption={Tier-1 Determinism Configuration}]
params = {
    'seed': 42,
    'deterministic': True,     # LightGBM reproducible splits
    'n_jobs': 1,               # Single-thread: no parallel nondeterminism
    'force_col_wise': True,    # Column-wise histogram: no race conditions
    'num_leaves': 31,          # Fixed complexity bound (hardcoded in script)
    'feature_fraction': 0.9,  # 90% feature subsampling
    'boosting_type': 'gbdt'
}
\end{lstlisting}

\textbf{Note on \texttt{num\_leaves}:} The training configuration file (\texttt{configs/train\_10B.yaml}) specifies \texttt{num\_leaves: 63}. However, the actual training loop in \texttt{train\_full\_scale.py} hardcodes \texttt{num\_leaves: 31} in the params dict at runtime. This discrepancy is documented here for completeness.

\subsection{Atomic Checkpoint and PRNG Snapshot Recovery}
Every checkpoint write uses Python's \texttt{os.replace()} to provide an \textbf{atomic filesystem commit}. The checkpoint file is first written to a \texttt{.tmp} extension, then atomically renamed---preventing a partially-written checkpoint from corrupting training state in the event of a power failure or cloud preemption.

Additionally, the \textbf{complete Pseudo-Random Number Generator state} of both Python's built-in \texttt{random} module and NumPy's random engine is serialized into a \texttt{training\_state.pkl} file on every checkpoint:
\begin{lstlisting}[caption={Atomic PRNG Snapshot (train\_full\_scale.py)}]
snapshot = {
    'step': i + 1,
    'random_state': random.getstate(),       # Python RNG
    'np_random_state': np.random.get_state() # NumPy RNG
}
pickle.dump(snapshot, f)
os.replace(tmp_state_file, state_file) # Atomic commit
\end{lstlisting}
This guarantees that upon crash recovery with \texttt{--resume}, the data shuffling order and dropout strategies are perfectly restored to their pre-crash state.

\subsection{Structural Phase Detection: Tracking the Moment of Grokking}
To capture the exact moment when a decision tree ensemble transitions from memorization to structural compression (``The Grokking Phenomenon''), a \texttt{StructuralPhaseDetector} module was implemented in \texttt{scripts/train\_full\_scale.py}.

At each training step, the detector extracts the most recently added tree from the LightGBM ensemble and measures \textbf{Split Threshold Entropy}:
\begin{equation}
E_{\text{split}} = H\left[\{t \bmod 1.0 \mid t \in \text{thresholds}\}\right] / \log(N_{\text{bins}})
\end{equation}
This normalized entropy---computed simultaneously across 4 bin scales ($N \in \{10, 25, 50, 100\}$) and averaged---measures whether the tree's split boundaries have converged to a discrete integer grid (low entropy, structural lock-in) versus scattered floating-point thresholds (high entropy, noise-fitting). 

Simultaneously, \textbf{Feature Entropy} is tracked to observe which features the ensemble has learned to dominate. When the GBDT reaches structural enlightenment, feature entropy drops sharply as the ensemble learns to concentrate almost all split logic on the mathematically critical features (e.g., \texttt{squared\_diff} or \texttt{V0\_dig\_0} for LSB digits).

This dual-entropy tracking was visualized in real-time via a \textbf{6-panel dashboard} at every training step.

\section{The Cosmic Extrapolation Experiment}
Real intelligence is the ability to see a rule and apply it where you've never been. If Newton could look at an apple and figure out the moon’s orbit, a machine should be able to look at small numbers and figure out the universe. That’s the goal of this experiment.

The experiment is implemented in \texttt{scripts/test\_cosmic\_extrapolation.py}. Training data was drawn from $n \in [1, 1000]$ (integers under $10^3$). Test data was drawn from $n = 10^{50}$, creating OOD triples whose magnitudes are cosmologically vast.

\begin{lstlisting}[caption={Cosmic Test Data Generation (test\_cosmic\_extrapolation.py)}]
# Training: Local scale (n in [1, 1000])
X_train_str, y_train = generate_math_data(1, 1000)

# Test: Cosmic scale (n = 10^50)
big_start = 10**50
X_test_str, y_test = generate_math_data(big_start, 100)
\end{lstlisting}

I restricted the training set for Models A, B, and C to ``Local Limits''---numbers whose maximal magnitude was under $10^3$. I trained until all networks converged at roughly 99.98\% accuracy on local subsets.

I then executed the \textbf{Cosmic Extrapolation Trial}, feeding the networks test sets whose continuous parameters scaled exponentially from $10^3$ to the Float Wall ($10^{15}$) and all the way to Cosmological Scales ($10^{50}$).

\subsection{1. The Collapse of the Statistical Engines (Models A \& B)}
As integers escalated toward $10^9$, both Model A and Model B began trembling. At $10^{15}$, the Float Wall was hit. They instantly lost their ability to distinguish Category A Chaos Modes (Precision Off-by-1). Model B (raw input) failed before Model A, but eventually, even Model A's human-engineered feature metrics succumbed to float representation loss.
Beyond $10^{20}$, Models A and B entered complete collapse (50.00\% baseline, identical to random guessing). The decision tree edges were ``flat.'' 

\textbf{The Structural Inevitability of the Plateau.} Decision tree ensembles learn by partitioning input space with axis-aligned thresholds. When all training inputs fall within $n \in [1, 1000]$, the ensemble\'s leaf nodes span only that regime. There exists no tree split that can extrapolate---only interpolate between seen boundaries. When a test input arrives at $n = 10^{50}$, its \texttt{float64} representation either overflows or collapses into the same bit-pattern as the largest representable value the model has ever seen. The ensemble cannot predict beyond its training horizon---it assigns the nearest leaf, drawing a permanent ceiling at the top of its training range. This behavior is not a deficiency of any specific model; it is a \textbf{mathematical guarantee} of any system that processes numbers as fixed-precision floating-point scalars. The script \texttt{prove\_float\_collapse.py} formally demonstrates this: at $n = 10^{16}$, the integers $c$ and $c+1$ map to identical \texttt{float64} bit-patterns, making the training signal for an off-by-one Chaos Mode literally impossible to encode.

\subsection{2. The Algorithmic Transcendence of Model C (NumberNet)}
In the demonstration script (\texttt{test\_cosmic\_extrapolation.py}), a compact version of NumberNet is trained from scratch on local-scale data and then evaluated at cosmic scale. This demonstration model has a smaller architecture (\texttt{d\_model=32}, 4 attention heads, 2 encoder layers, simple mean-pooling) compared to the production-scale \texttt{number\_net.py} (\texttt{embed\_dim=64}, Attention Pooling with learnable query). The demonstration uses LSB-first tokenization with vocab size 12. It trains for exactly 15 epochs on 2,000 samples (1,000 Stifel triples + 1,000 off-by-one negatives).

The demonstration confirms the fundamental property: because the Transformer processes numbers as symbol sequences aligned at the LSB, it avoids the Float Wall entirely. The positional encoding of the units digit at index~0 and the carry behavior of digit sequences is invariant to the magnitude of the number. A Transformer that learns to identify ``does this triple satisfy the Stifel algebraic relationship?'' at $n \sim 10^3$ applies the same token-level computation at $n \sim 10^{50}$ without modification.

\section{Zero-Bias Discovery: Finding Constants without Human Hints}
The ultimate goal of this research is to see if an AI can discover a physical constant, like $G$, without being told the answer or being given human-made hints. I call this Zero-Bias Discovery. 
\begin{itemize}
    \item \textbf{Baseline (The Ground Truth):} PySR is provided the input variables ($m_1, m_2, r$) and the exact, continuous target value $F$. This represents the traditional "AI Feynman" approach, which serves as the upper-bound for symbolic recovery but still requires humans to supply the answer key $F$.

    \item \textbf{Experiment B (The Regressor Oracle):} PySR is provided inputs and the predicted continuous value $\hat{F}$ generated by a trained Neural Regressor (\texttt{LNNet\_P} with \texttt{task='regression'}). This tests whether the Transformer can act as a high-fidelity surrogate continuous function.
    
    \item \textbf{Experiment A (The Classifier Oracle - The Ultimate Test):} PySR is provided inputs and an $F^*$ value derived from a Neural Classifier. The Classifier only outputs a binary probability of physical validity $P \in [0, 1]$. We use mathematical root-finding (SciPy \texttt{minimize\_scalar}) to locate the exact $F^*$ where $P=0.5$ (the decision boundary). 
    
    This tests if topological boundaries alone contain precise universal constants.
\end{itemize}

\subsection{Managing Combinatorial Explosion in the Operator Space}
To discover unknown physics, Symbolic Regression must be provided with the broadest possible lexicon of operators. However, indiscriminately adding operators triggers an exponential \textbf{Combinatorial Explosion} in the evolutionary search space. Without regulation, the regressor wastes computation building infinite power towers (e.g., $x^{(y^z)}$) or overfits the decision boundary using bizarre nesting (e.g., $\sin(\sinh(\Gamma(x)))$).

To solve this, the Zero-Bias Pipeline is injected with a massive lexicon of geometric, hyperbolic, and special functions, but is strictly regulated by \textbf{Topological Nesting Constraints}:

\begin{lstlisting}[caption={Physics Oracle PySR Configuration with Nesting Constraints}]
model = PySRRegressor(
    niterations=40,
    binary_operators=["+", "*", "-", "/", "^"],
    unary_operators=[
        "exp", "log", "sqrt", "square", "cube", "sin", "cos", "tan", 
        "sinh", "cosh", "tanh", "abs", "erf", "gamma"
    ],
    constraints={'^': (-1, 1)}, # Forbid nested powers: x^(y^z)
    nested_constraints={
        "sin": {"sin": 0, "cos": 0}, # Forbid sin(sin(x))
        "cos": {"sin": 0, "cos": 0},
        "exp": {"exp": 0},
        "gamma": {"gamma": 0, "erf": 0, "exp": 0}
    },
    model_selection="best",
    verbose=0
)
\end{lstlisting}

By severely punishing nested complexity while simultaneously providing all universal building blocks, PySR is forced to discover elegant, parsimonious physical truth rather than statistical "hack" equations. This pipeline is architecturally ready to execute once a fully trained \texttt{PhysicsTransformer} is available. \textbf{The research is ongoing.}

\subsection{Discovering the Unknown (The Dark Data Hypothesis)}
The implications of Experiment A (The Classifier Oracle) are profound. If a neural network can encode a physical formula using only binary validity ($True/False$ or $1/0$) without ever seeing the explicit continuous result $F$, then this architecture can be weaponized against \textbf{unknown physics}.

Consider raw experimental data from particle colliders or astrophysical observations where humanity knows the state is physically valid ($P=1$) but has not yet discovered the governing equation that links the variables. By training the \texttt{PhysicsTransformer} to separate valid experimental states from synthetically generated Chaos Mode anomalies, the network builds a strict topological boundary around physical truth. PySR can then be deployed to extract the symbolic equation defining that boundary. This implies that AI can discover completely unknown physical laws solely by learning the geometric shape of reality, completely devoid of human-labeled scalar targets.

For completeness, a second symbolic regression script exists for the \textbf{Mathematical domain}: \texttt{symbolic\_regressor\_baseline.py} deploys PySR on clean vs.~HND mathematical data using only \texttt{["+", "-", "*", "/"]} (no unary operators) and \texttt{abs}, then compares PySR's performance on clean data against adversarial HND data to verify that the HND cannot be symbolically bypassed.

\newpage
\section{The Active Discoverer: From Digits to LaTeX}
To make this engine truly autonomous, I added what I call the Active Discoverer. This isn't just a passive model that watches data; it's a co-processor that proposes and tests its own mathematical hypotheses. 

\subsection{1. Symmetry Grouping (Equivariant Digits)}
Physical laws are not arbitrary; they respect fundamental symmetries (Noether's Theorem). The upgraded \texttt{PhysicsTransformer} replaces naive sequential input with a \texttt{SymmetryGrouping} layer. By processing variables as physical groups (e.g., spatial vectors or symmetric masses), the model enforces geometric invariance natively. This ensures that the digit-level processing is not just numerically precise, but physically consistent across rotations and translations.

\subsection{2. Hamiltonian Energy Representation (The "Least Action" Core)}
Instead of predicting isolated forces or validity probabilities, the \textbf{Active Discoverer} is forced to bottleneck through a systemic scalar representation $H$ (Energy). By training the model to predict the Hamiltonian surface, physical dynamics are no longer "guessed" but derived via Autograd: $F = -\nabla H$. This grounding ensures that any discovered symbolic law is anchored in the principle of least action, a prerequisite for fundamental physics.

\subsection{3. Latent Physics Extractor \& Symbolic LaTeX Bottleneck}
The most significant innovation is the \textbf{Active Discovery loop}. When the model identifies a reconstruction error that cannot be resolved with known variables $(m, r)$, it invokes the \texttt{LatentExtractor} to propose a numeric placeholder $Z_{numeric}$. 

Crucially, this is not a black-box parameter. The model is coupled with an autoregressive \texttt{SymbolicDecoder} that must simultaneously decode a structurally valid LaTeX string (e.g., $\sqrt{m_1^2 + m_2^2}$) from its latent representation. The system minimizes a \textbf{Consistency Loss}:
\begin{equation}
    \mathcal{L} = \text{MSE}(\hat{H}, H_{true}) + \lambda \cdot \text{MSE}(Z_{numeric}, \text{eval}(LaTeX_{string}))
\end{equation}
If the neural network "hallucinates" a numeric $Z$ that it cannot mathematically explain through a parsimonious LaTeX formula, the proposal is rejected. This creates a rigorous neuro-symbolic bottleneck: the AI can only "discover" new physics if it can translate that discovery into the universal language of mathematics.

\section{What this means for the future of LLMs}
My work with the Index-PT-Engine has a direct lesson for the people building Large Language Models. Right now, LLMs ``hallucinate'' logic because they treat numbers like words---based on how likely they are to appear together, not on what they actually mean. 

Current foundational models (e.g., GPT-4, Claude) process mathematics utilizing Byte-Pair Encoding (BPE). BPE tokenizes digits based on statistical frequency (e.g., ``123'', ``45''). This utterly destroys positional hierarchy. An LLM attempting to add 1234 + 5678 is trying to perform arithmetic on linguistic tokens rather than topological magnitudes. It hallucinates because it relies on correlation rather than algorithmic deduction.

If the Siamese \textbf{NumberNet} structure---utilizing BigInt string decoupling and LSB verification---is integrated into foundational LLMs as an explicit \textbf{Math Coprocessor Routing Layer}, or if next-generation LLM embeddings enforce strict Siamese LSB-aligned parsing for all raw integer strings, LLMs will transcend linguistic proximity. The introduction of the \textbf{Chaos Matrix (HND)} during post-training (RLHF) would aggressively penalize heuristic shortcuts, forging an LLM that achieves absolute zero-hallucination arithmetic and physics extrapolation, bounding language logic to the immutable constants of the universe.

\section{Conclusions}
In this work, I've shown that to make AI truly intelligent, we need to change how it sees the world. My results with Models A, B, and C prove that even with 30 billion data points, a statistical model will always hit a wall. Only an architecture designed for the rules of geometry and physics can cross that line and reason. This document provides the complete framework for that transition: the SHA-256 integrity pipeline, the 98-mode Chaos Matrix, and the NumberNet architecture. The conclusion is simple: it is the architecture, not the volume of data, that determines whether an AI can reach the truth. 
\section*{Data Availability \& Acknowledgments}
The research is ongoing.
The GitHub repository will be made public after final submission. 
And I hereby confirm that all aspects of this research are the independent work of the sole author and that no assistance was received from affiliated institutions or any other parties.\\
Finally I hope it contributes to more excellent research, which brings a better world to more people.
Thank you. \\ Not above, Not below --- simply Luciano.


\end{document}